\documentclass[10pt,twocolumn,letterpaper]{article}

\usepackage{iccv}
\usepackage{times}
\usepackage{epsfig}
\usepackage{graphicx}
\usepackage{amsmath}
\usepackage{amssymb}
\usepackage{booktabs}
\usepackage{algorithm}
\usepackage{algpseudocode}
\usepackage{mathtools}
\usepackage{multirow}
\usepackage{color}
\usepackage{xcolor}
\usepackage{capt-of}

\usepackage[sort,nocompress]{cite}

\usepackage[pagebackref=true,breaklinks=true,letterpaper=true,colorlinks,bookmarks=false]{hyperref}

\DeclareMathOperator*{\argmin}{arg\,min}

\iccvfinalcopy 

\def\iccvPaperID{2015} 

\begin{document}

\title{Guided Motion Diffusion for Controllable Human Motion Synthesis}


\newcommand*{\affaddr}[1]{#1}
\newcommand*{\affmark}[1][*]{\textsuperscript{#1}}
\newcommand*{\email}[1]{\small{\texttt{#1}}}
\author{
Korrawe Karunratanakul\affmark[1] \quad
Konpat Preechakul\affmark[2] \quad
Supasorn Suwajanakorn\affmark[2] \quad
Siyu Tang\affmark[1] 
\vspace{0.7em} \\
\affmark[1]{ETH Z{\"u}rich, Switzerland} \quad
\affmark[2]{VISTEC, Thailand}  \\
{\small\url{https://korrawe.github.io/gmd-project/}}  \vspace{-0.5em}
}


\newcommand{\myparagraph}[1]{\vspace{0.5em}\noindent\textbf{#1}}
\newcommand{\ST}[1]{{\color{red}[ST: #1]}}
\newcommand{\KK}[1]{{\color{magenta}[KK: #1]}} 
\newcommand{\TODO}[1]{{\color{blue}[TODO: #1]}} 
\newcommand{\KP}[1]{{\color{blue}[KP: #1]}}
\newcommand{\kp}[1]{\KP{#1}}
\newcommand{\st}[1]{\ST{#1}}
\newcommand{\todo}[1]{{\color{RedOrange}\textbf{TODO}: #1}}
\newcommand{\V}[1]{\mathbf{#1}}
\newcommand{\R}[0]{\rm I\!R}
\newcommand{\E}[0]{\rm I\!E}
\newcommand{\loss}[0]{\mathcal{L}}

\newcommand{\methodname}{{GMD}}
\newcommand{\methodnamefull}{{???~}}

\newcommand{\vaename}{{???~}}

\newcommand{\norm}[1]{\left\lVert#1\right\rVert}

\newcommand{\cmark}{\ding{51}}%
\newcommand{\xmark}{\ding{55}}%

\newcommand{\xt}{{\mathbf{x}_t}}
\newcommand{\xtone}{{\mathbf{x}_{t-1}}}
\newcommand{\xzero}{{\mathbf{x}_0}}
\newcommand{\x}{\mathbf{x}}
\newcommand{\y}{\mathbf{y}}
\newcommand{\m}{\mathbf{m}}
\newcommand{\z}{\mathbf{z}}
\newcommand{\I}{\mathbf{I}}
\newcommand{\Pyx}{P_y^x}
\newcommand{\Pxy}{P_x^y}
\newcommand{\Myx}{M_y^x}
\newcommand{\Pxz}{P_x^z}
\newcommand{\Pzx}{P_z^x}
\newcommand{\Mzx}{M_z^x}
\newcommand{\Pzy}{P_z^y}
\newcommand{\Pyz}{P_y^z}
\newcommand{\Myz}{M_y^z}
\newcommand{\Mzz}{M_{z^*}^z}
\newcommand{\defeq}{\vcentcolon=}
\newcommand{\Gx}{G_x}
\newcommand{\Gz}{G_z}

\twocolumn[{%
\renewcommand\twocolumn[1][]{#1}%
\maketitle

\begin{center}
  \newcommand{\teaserwidth}{\textwidth}
  \vspace{-0.3cm}
  \centerline{\includegraphics[width=\linewidth]{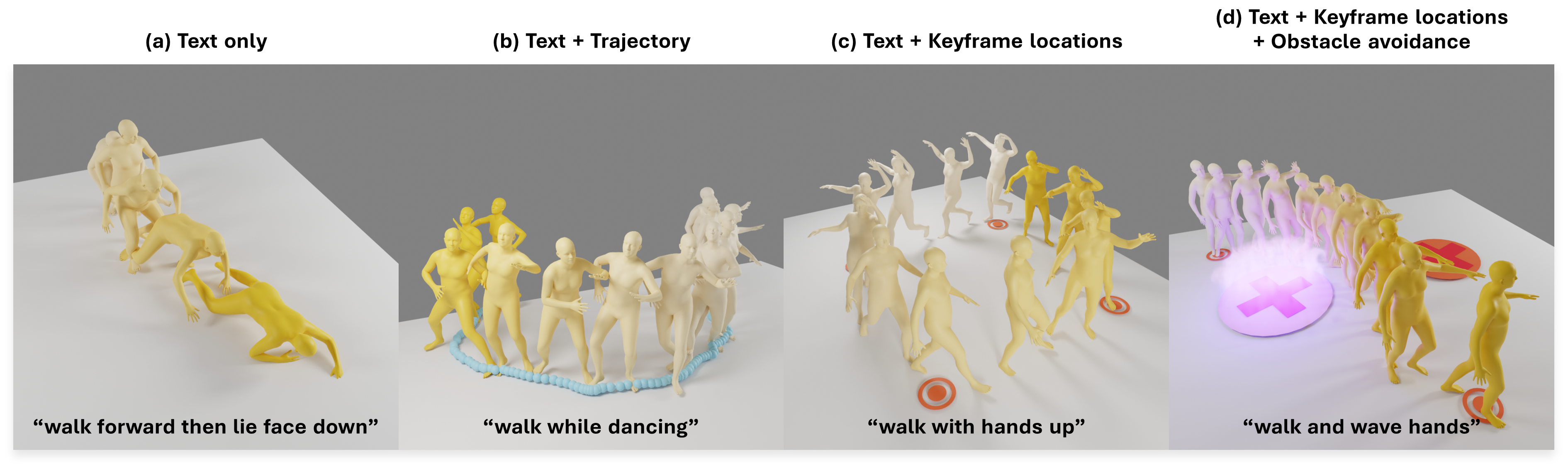}}
    \captionof{figure}{Our proposed Guided Motion Diffusion (\textbf{GMD}) can generate high-quality and diverse motions given a text prompt and a goal function. We demonstrate the controllability of GMD on four different tasks, guided by the following conditions: (a) text only, (b) text and trajectory, (c) text and keyframe locations (double circles), and (d) with obstacle avoidance (red-cross areas represent obstacles). The darker the colors, the later in time. 
    }
    \label{fig:teaser}
\end{center}%
}]

\begin{abstract}
    \vspace{-0.5cm}
   Denoising diffusion models have shown great promise in human motion synthesis conditioned on natural language descriptions. However, integrating spatial constraints, such as pre-defined motion trajectories and obstacles, remains a challenge despite being essential for bridging the gap between isolated human motion and its surrounding environment. 
   To address this issue, we propose Guided Motion Diffusion (\textbf{GMD}), a method that 
   incorporates spatial constraints into the motion generation process. 
   Specifically, we propose an effective feature projection scheme that manipulates motion representation to enhance the coherency between spatial information and local poses. Together with a new imputation formulation, the generated motion can reliably conform to spatial constraints such as global motion trajectories. Furthermore, given sparse spatial constraints (e.g.~sparse keyframes), we introduce a new dense guidance approach to turn a sparse signal, which is susceptible to being ignored during the reverse steps, into denser signals to guide the generated motion to the given constraints. 
   Our extensive experiments justify the development of \methodname,
   which achieves a significant improvement over state-of-the-art methods in text-based motion generation while allowing control of the synthesized motions with spatial constraints.
\end{abstract}


\vspace{-1.1cm}
\section{Introduction}
Recently, denoising diffusion models have emerged as a promising approach for human motion generation \cite{zhao2023modiff,zhou2022ude,dabral2022mofusion} outperforming other alternatives such as GAN or VAE in terms of both quality and diversity \cite{Tevet2022-ih,chen2022mld,yuan2022physdiff}.
Several studies have focused on generating motion based on expressive text prompts \cite{Tevet2022-ih,chen2022mld}, or music \cite{zhou2022ude,tseng2022edge}.
The state-of-the-art motion generation methods, such as MDM\cite{Tevet2022-ih}, utilize classifier-free guidance to generate motion conditioned on text prompts.
However, incorporating spatial constraints into diffusion models remains underexplored.
Human motions consist of both semantic and spatial information, where the semantic aspect can be described using natural languages or action labels and the spatial aspect governs physical interaction with surroundings. To generate realistic human motion in a 3D environment, both aspects must be incorporated.
Our experiments show that simply adding spatial constraint guidance, such as global trajectories, into the state-of-the-art models or using imputation and in-painting approaches do not yield satisfactory results. 

\begin{figure}[h]
    \centering
    \includegraphics[width=\linewidth]{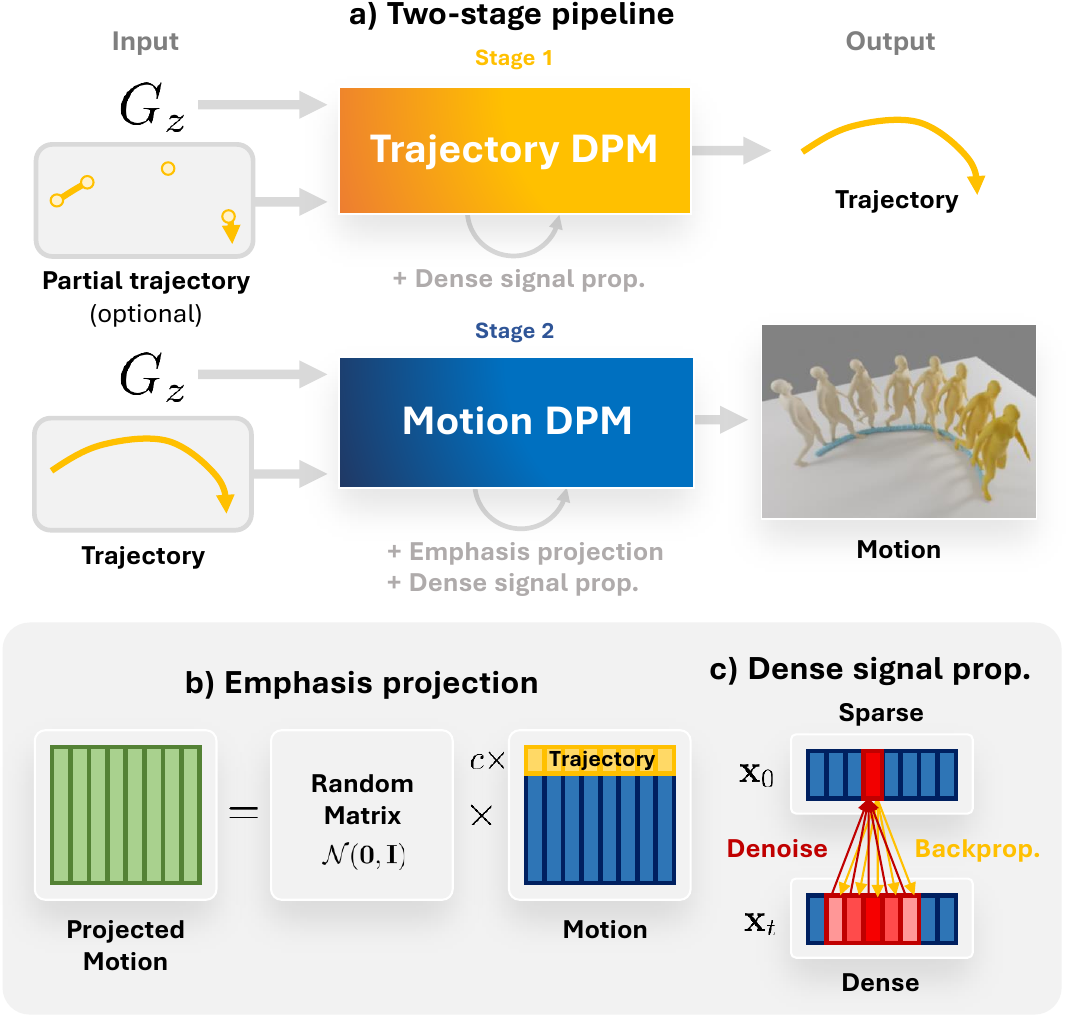}
    \caption{We tackle the problem of spatially conditioned motion generation with \methodname, depicted in \textbf{a)}. Our main contributions are \textbf{b) Emphasis projection}, for better trajectory-motion coherence, and \textbf{c) Dense signal propagation}, for a more controllable generation even under sparse guidance signal.}
    \label{fig:pipeline}
\end{figure}

We identify two main issues that make the motion diffusion models likely to ignore the guidance when conditioned on spatial objectives: the sparseness of global orientation in the motion representation and sparse frame-wise guidance signals.
By design, the diffusion models are a denoising model that consecutively denoises the target output over multiple steps. With sparse guidance, a small portion of the output that receives guidance will be inconsistent with all other parts that do not, therefore, are more likely to be treated as noise and discarded in subsequent steps.

First, the sparseness within a frame is a result of common motion representations that separate local pose information, like joint rotations, from global orientations, such as pelvis translations and rotations \cite{rempe2021humor}, usually  with more focus on local poses.
For instance, the common motion representation \cite{guo2022t2m} uses 4 values to represent global orientation and 259 values for local pose in each frame.
Such imbalance can cause the model to focus excessively on local pose information, and consequently, perceive guided global orientation as noise, resulting in a discrepancy such as foot skating.

Second, in many applications such as character animation, gaming, and virtual reality, the spatial control signals are defined on only a few keyframes such as target locations on the ground.
We show that the current diffusion-based motion generation models struggle to follow such sparse guidance as doing so is equivalent to guiding an image diffusion model with only a few pixels. As a result, either the guidance at the provided keyframes will be ignored during the denoising process or the output motion will contain an artifact where the character warps to satisfy the guidance only in those specific keyframes.

To effectively incorporate sparse spatial constraints into the motion generation process, we propose GMD, a novel and principled Guided Motion Diffusion model.
To alleviate the discrepancy between local
pose and global orientation in the guided denoising steps, we introduce {\it emphasis projection}, a general representation manipulation method that we use to increase the importance of spatial information during training.
%
Additionally, we derive a new imputation and inpainting formulation that enables the existing inpainting techniques to operate in the projected space, which we leverage to generate significantly more coherent motion under guidance by spatial conditions.
Then, to address the highly sparse guidance, we draw inspiration from the credit assignment problem in Reinforcement Learning \cite{Sutton1988-dh,Watkins1989-ol}, where sparse rewards can be distributed along a trajectory to allow for efficient learning \cite{Arjona-Medina2019-sk}. 
%
Our key insight is that motion denoisers, including the diffusion model itself, can be used to expand the spatial guidance signal at a specific location to its neighboring locations without any additional model. By turning a sparse signal into a dense one by back-propagating through a denoiser, it enables us to achieve high-quality controllable motion synthesis, even with extremely sparse guidance signals.

In summary, our contributions are:
(1) Emphasis projection, a method to adjust relative importance between different parts of the representation vector, which we use to encourage coherency between spatial information and local poses to allow spatial guidance.
(2) Dense signal propagation, a conditioning method to tackle the sparse guidance problem. 
(3) \methodname, an effective spatially controllable motion generation method that enables the unexplored synthesizing of motions based on free-text and spatial conditioning by integrating the above contributions into our proposed Unet-based architecture.
We provide extensive analysis to support our design decisions and show the versatility of \methodname~on three tasks: trajectory conditioning, keyframe conditioning, and obstacle avoidance.
Additionally, \methodname's model also significantly outperforms the state-of-the-art in traditional text-to-motion tasks.

\section{Related Work}
\myparagraph{Diffusion-based probabilistic generative models (DPM).}
DPMs \cite{Ho2020-ew, Sohl-Dickstein2015-hp, Song2019-xr, Song2020-yj} have gained significant attention in recent years due to their impressive performance across multiple fields of research. They have been used for tasks such as image generation \cite{Dhariwal2021-bt}, image super-resolution \cite{Saharia2021-lm, Li2021-qr}, speech synthesis \cite{Kong2020-sn, Kong2020-sn, Popov2021-qe}, video generation \cite{Ho2022-qm, Ho2022-gh}, 3D shape generation \cite{Poole2022-kx, Watson2022-vm}, and reinforcement learning \cite{Janner2022-no}.

The surge in interest in DPMs may be attributed to their impressive controllable generation capabilities, including text-conditioned generation \cite{Ramesh2022-id,Rombach2021-nu,Saharia2022-ns} and image editing \cite{Meng2021-bn,Choi2021-oe,Brooks2022-wx,Hertz2022-td,Balaji2022-gh}.
Latent diffusion models (LDM) are another area of interest, which includes representation learning \cite{Preechakul2022-ql, Kwon2023-bu} and more efficient modeling techniques \cite{Rombach2021-nu,chen2022mld}. 

Moreover, DPMs exhibit a high degree of versatility in terms of conditioning. There are various methods for conditional generation, such as imputation and inpainting \cite{Chung2022-tm,Choi2021-oe,Meng2021-bn}, classifier guidance \cite{Dhariwal2021-bt, Chung2022-tm}, and classifier-free guidance \cite{Rombach2021-nu,Saharia2022-ns,Ramesh2022-id, ho2021classifierfree}. Inpainting and classifier guidance can be applied to any pretrained DPM, which extends the model's capabilities further without the need for retraining.

\myparagraph{Human motion generation.}
The goal of the human motion generation task is to generate motions based on the conditioning signals. 
Various conditions have been explored such as partial poses \cite{duan2021single,harvey2020robust,Tevet2022-ih}, trajectories \cite{wang2021synthesizing,zhang2021learning,kaufmann2020convolutional}, images \cite{rempe2021humor,chen2022learning}, music \cite{lee2019dancing,li2022danceformer,li2021ai}, text \cite{ahuja2019language2pose,guo2022t2m,guo2022tm2t,kim2022flame,petrovich2022temos}, objects \cite{wu2022saga}, action labels \cite{petrovich2021action,guo2020action2motion}, or unconditioned \cite{pavlakos2019expressive,yan2019convolutional,zhang2020perpetual,zhao2020bayesian}.
Recently, many diffusion-based motion generation models have been proposed \cite{ma2022mofusion,kim2022flame,zhao2023modiff,zhou2022ude,dabral2022mofusion} and demonstrate better quality compared to alternative models such as GAN or VAE.
Employing the CLIP model \cite{radford2021learning}, these models showed great improvements in the challenging text-to-motion generation task \cite{chen2022mld,yuan2022physdiff,zhang2022motiondiffuse} as well as allowing conditioning on partial motions \cite{Tevet2022-ih} or music \cite{tseng2022edge,alexanderson2022listen}. However, they do not support conditioning signals that are not specifically trained, for example, following keyframe locations or avoiding obstacles.
Maintaining the capabilities of the diffusion models, we propose methods to enable spatial guidance without retraining the model for each new objective.

\section{Background}

\subsection{Diffusion-based generative models} 

Diffusion-based probabilistic generative models (DPMs) are a family of generative models that learn a sequential denoising process of an input $\xt$ with varying noise levels $t$. The noising process of DPM is defined cumulatively as $q(\xt|\xzero) = \mathcal{N}(\sqrt{\alpha_t}\xzero, (1 - \alpha_t) \I)$, where $\xzero$ is the clean input, $\alpha_t = \prod_{s=1}^t (1-\beta_s)$, and $\beta_t$ is a noise scheduler. The denoising model $p_\theta(\xtone|\xt)$ with parameters $\theta$ learns to reverse the noising process by modeling the Gaussian posterior distribution $q(\xtone|\xt,\xzero)$. DPMs can map a prior distribution $\mathcal{N}(\mathbf{0}, \mathbf{I})$ to any distribution $p(\x)$ after $T$ successive denoising steps.

To draw samples from a DPM, we start from a sample $\x_{T}$ from the prior distribution $\mathcal{N}(\mathbf{0},\mathbf{I})$. 
Then, for each $t$, we sample $\xtone \sim \mathcal{N}(\mathbf{\mu}_{t}, \Sigma_t)$ until $t = 0$, where 
\begin{equation}\label{eq:sampling}
\mathbf{\mu}_{t} = \frac{\sqrt{\alpha_{t-1}} \beta_t}{1-\alpha_t}\x_{0} + \frac{\sqrt{1-\beta_t}(1-\alpha_{t-1})}{1-\alpha_t}\xt    
\end{equation}
and $\Sigma_t$ is a variance scheduler of choice, usually $\Sigma_t = \frac{1-\alpha_{t-1}}{1-\alpha_t}\beta_t$ \cite{Ho2020-ew}. $\x_{0}$ in Eq.~\ref{eq:sampling} is the prediction from a denoising model. For an $\epsilon_\theta$ model, $\x_{0}= \frac{1}{\sqrt{\alpha_t}} \xt + \frac{\sqrt{1 - \alpha_t}}{\sqrt{\alpha_t}} \epsilon_\theta(\xt)$.

There are multiple choices for the denoising model to predict including the clean input $\xzero$, the noise $\epsilon$, and the one-step denoised target $\mu_t$. 
An $\x_{0,\theta}$ model is trained using the squared loss to the clean input $\norm{\x_{0,\theta}(\xt) - \xzero}^2$, an $\epsilon_\theta$ model is trained using the squared loss $\norm{\epsilon_\theta(\xt) - \epsilon}^2$, and $\mu_{t,\theta}$ model is trained using the squared loss $\norm{\mu_{t,\theta}(\xt) - \mu_t}^2$  .

\subsection{Controllable generation with diffusion models.}\label{sec:dpm_cond}

\myparagraph{Classifier-free guidance.}
The conditioning signals are treated as additional inputs to the denoiser $p_\theta(\xtone|\xt, d)$ where $d$ is the conditioning signals which can be omitted $d = \varnothing$ to generate unconditionally. Classifier-free guidance has been shown to generate very high-quality results \cite{Saharia2022-ns, Rombach2021-nu, Tevet2022-ih}. To draw samples, the effective denoiser becomes  $\hat{\epsilon}_\theta(\xt, d) = w \epsilon_\theta(\xt, d) + (1-w) \epsilon_\theta(\xt, \varnothing)$, where $w$ controls the conditional strength.
The new $\hat{\epsilon}_\theta$ model can be used in Eq.~\ref{eq:sampling}.
Two downsides of this method are that the nature of conditioning signals need to be known before hand and the denoiser needs to be adjusted and retrained for each specific case restricting its flexibility. 

\myparagraph{Classifier guidance.}
We can also obtain $p(\xtone|\xt, d)$ from $p_\theta(\xtone|\xt)p(d|\xt)$ \cite{Dhariwal2021-bt} , where $p(d|\xt)$ is any probability function that we can approximate its score function $\nabla_\xt \log p(d|\xt)$ effectively. 
The new sampling process is similar to the original (Eq.~\ref{eq:sampling}) but with the mean shifted by the scaled score function as
\begin{equation}\label{eq:class_guidance}
    \mathbf{\mu}_{t} = \mathbf{\mu}'_{t} + s \Sigma_t \nabla_\xt \log p(d|\xt)
\end{equation}
where $\mathbf{\mu}'_{t}$ is the original mean, $s$ controls the conditioning strength, and $\Sigma_t$ is a variance scheduler which can be the same as in Eq.~\ref{eq:sampling}. 
Since $\Sigma_t$ is a decreasing sequence, the guidance signal diminishes as $t \rightarrow 0$ which corresponds to the characteristic of DPMs that tend to modify $\xt$ less and less as time goes.
Classifier guidance is a post-hoc method, i.e., there is no change to the DPM model, one only needs to come up with $p(d|\xt)$ which is extremely flexible. 

\myparagraph{Imputation and inpainting.}\label{sec:imputation} 
To generate human motion sequences from partial observations, such as global motion trajectories or keyframe locations, inpainting is used. These partial observations, called imputing signals, are used to adjust the generative process towards the observations. Imputation and inpainting are two sides of the same coin.

Let $\y$ be a partial target value in an input $\x$ that we want to impute. The imputation region of $\y$ on $\x$ is denoted by $\Myx$, and a projection $\Pyx$ that resizes $\y$ to that of $\x$ by filling in zeros.
In DPMs, imputation can be done on the sample $\xtone$ after every denoising step \cite{Chung2022-tm}. We have the new imputed sample $\tilde{\x}_{t-1}$ as
\begin{equation}\label{eq:imputation}
\tilde{\x}_{t-1} = (1 - \Myx) \odot \x_{t-1} + \Myx \odot \Pyx \y_{t-1}
\end{equation}
where $\odot$ is a Hadamard product and $\y_{t-1}$ is a noised target value. $\y_{t-1} \sim \mathcal{N}(\sqrt{\alpha_{t-1}} \y, (1 - \alpha_{t-1}) \mathbf{I})$ following Ho~\etal~\cite{Ho2020-ew} is one of the simplest choices of $\y_{t-1}$. 

Note that all three modes of conditioning presented here are not mutually exclusive. One could apply one or more tricks in a single pipeline.

\section{Guided Motion Diffusion}

\newcommand{\comment}[1]{\textcolor{gray}{#1}}

\begin{algorithm}
\caption{\textbf{GMD}'s two-stage guided motion diffusion}\label{alg:key_loc}
\begin{algorithmic}[1]
\Require A trajectory DPM $\z_{0, \phi}$, a motion DPM $\x_{0, \theta}$, a goal function $G_\z(\cdot)$, and keyframe locations $\y$ (if any).
\State \comment{\small \# Stage 1: Trajectory generation}
\State $\z_T \sim \mathcal{N}(\mathbf{0}, \mathbf{I})$
\ForAll{$t$ from $T$ to $1$}    
    \State $\z_0 \leftarrow \z_{0,\phi}(\z_t)$
    \State $\mu, \Sigma \leftarrow \mu(\z_0, \z_t), \Sigma_t $
    \State \comment{\small \# Classifier guidance (Eq. \ref{eq:class_guidance})}
    \State \comment{\small \# \textbf{Dense signal propagation}}
    \State $\z_{t-1} \sim \mathcal{N}\left(\mu - s\Sigma\nabla_{\z_t} \Gz(\z_0), \Sigma\right)$
    \State \comment{\small \# Impute $\y$ on $\z$ (Eq. \ref{eq:imputation}) (if any)}
    \State $\z_{t-1} \leftarrow (1 - \Myz) \odot \z_{t-1} + \Myz \y_{t-1}$
\EndFor
\State \comment{\small \# Stage 2: Trajectory-conditioned motion generation}
\State $\x_T^\text{proj} \leftarrow \text{sample from } \mathcal{N}(\mathbf{0}, \mathbf{I})$
\ForAll{$t$ from $T$ to $1$}
    \State $M \leftarrow \Pzx \Myz$  \comment{\small \quad\# Imputation region of $\y$ on $\x$}
    \State $\x^\text{proj}_0 \leftarrow \x^\text{proj}_{0,\theta}(\x_t^\text{proj})$ \comment{\small \quad\# \textbf{Emphasis projection}}
    \State \comment{\small  \# Impute $\y$ on $\x^\text{proj}$ (Eq. \ref{eq:impute_proj})}
    \State $\tilde{\x}_0^\text{proj} \leftarrow A\left( (1 - M) \odot A^{-1} \x^\text{proj}_0 + M  \Pzx \y \right)$
    \State $\mu, \Sigma \leftarrow \mu(\tilde{\x}_0^\text{proj}, \x_t^\text{proj}), \Sigma_t $
    \State \comment{\small \# Masked classifier guidance (Eq. \ref{eq:class_guidance_mask_proj})}
    \State \comment{\small \# \textbf{Dense signal propagation}}
    \State $\Delta \leftarrow - s \Sigma A^{-1} \nabla_{\x_t^\text{proj}} \Gz\big(\Pxz A^{-1} \x^\text{proj}_0 \big)$
    \State $\mu \leftarrow \mu + A (1-M) \odot \Delta$ 
    \State $\x_{t-1} \sim \mathcal{N}(\mu, \Sigma)$
\EndFor\\
\Return $\z_0$
\end{algorithmic}
\end{algorithm}

We aim to generate realistic human motions that can be guided by spatial constraints, enabling the generated human motion to achieve specific goals, such as following a global trajectory, reaching certain locations, or avoiding obstacles. Although diffusion-based models have significantly improved text-to-motion modeling\cite{Tevet2022-ih,chen2022mld}, generating motions that achieve specific goals is still beyond the reach of the current models. 
Our work addresses this limitation and advances the state-of-the-art in human motion modeling.

We are interested in modeling a full-body human motion that satisfies a certain scalar goal function $\Gx(\cdot)$ that takes a motion representation $\x$ and measures how far the motion $\x$ is from the goal (lower is better).
More specifically, $\x \in \mathbb{R}^{N \times M}$ represents a sequence of human poses for $M$ motion steps, where $N$ is the dimension of human pose representations, e.g., $N = 263$ in the HumanML3D \cite{guo2022t2m} dataset. 
Let $X$ be the random variable associated with $\x$.
Our goal is to model the following conditional probability using a motion DPM
\begin{equation}\label{eq:goal}
    p\big(\x|\Gx(X) = 0\big)
\end{equation}
This can be extended to $p\big(\x|\Gx(X) = 0, d \big)$, where $d$ is any additional signal, such as text prompts. From now on, we omit $d$ to reduce clutter.

Many challenging tasks in motion modeling can be encapsulated within a goal function $\Gz$ that only depends on the trajectory $\z$ of the human motion, not the whole motion $\x$. 
Let us define $\z \in \mathbb{R}^{L \times M}$ to be the trajectory part of $\x$ with length $M$ and $L = 2$ describing the ground location of the pelvis of a human body. 
A particular location $\z^{(i)}$ at motion step $i$ describes the pelvis location of the human body on the ground plane. 
We define a projection $\Pxz$ that resizes $\x$ to match $\z$ by taking only the $\z$ part, and its reverse $\Pzx$ that resizes $\z$ to match $\x$ by filling in zeros. With this, our conditional probability becomes $p\big(\x|\Gz(\Pxz X) = 0\big)$.

\begin{figure*}[h]
    \centering
    \includegraphics[width=0.85\linewidth]{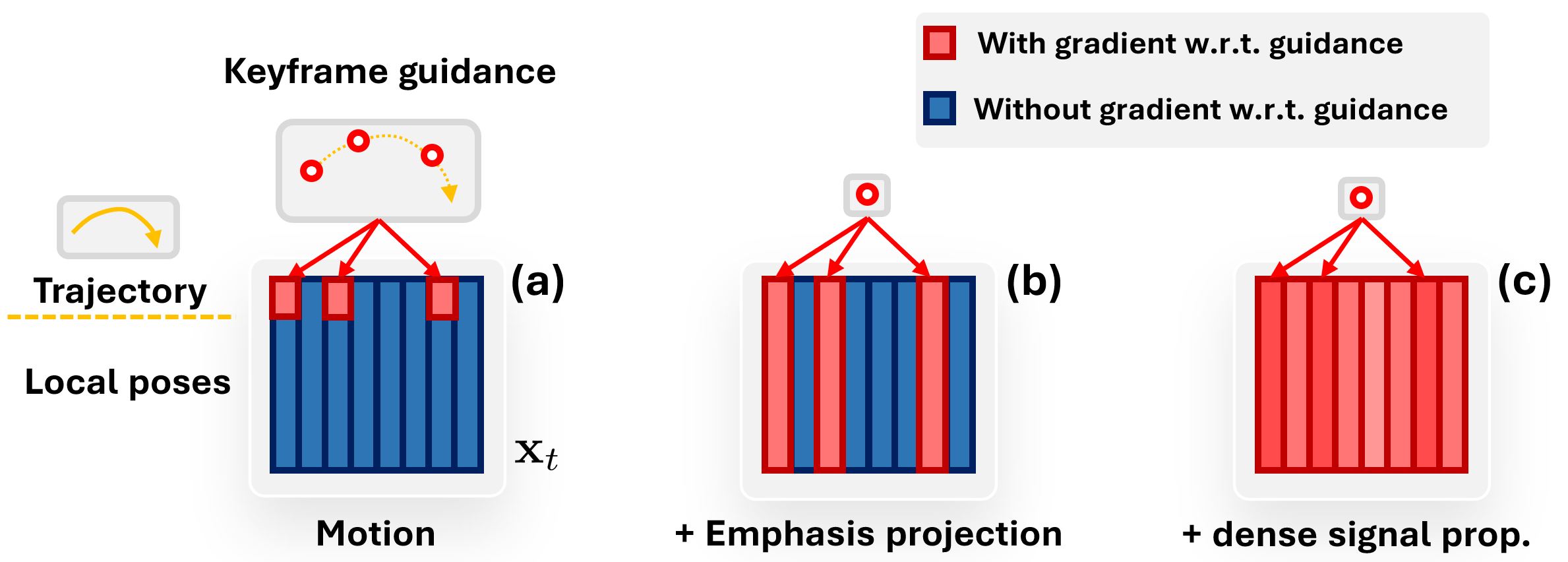}
    \caption{(a) Under standard motion representation and guiding method, only a few values in the motion representation are updated according to the guidance. (b) With Emphasis projection, all values in each frame describing the motion receives gradients w.r.t. the guidance, leading to better coherence between global orientation and local pose in each frame. (c) With dense gradient propagation, all frames are updated according to the guidance at the keyframes, making the guidance less likely to be ignored.}
    \label{fig:idea_overview}
\end{figure*}

In this work, we will show how text-to-motion DPMs can be extended to solve several challenging tasks, including trajectory-conditioned motion generation, location-conditioned trajectory planning, and obstacle avoidance trajectory planning. 
Using our proposed Emphasis projection and dense signal propagation, we alleviate the sparse guidance problem and enable motion generation based on spatial conditions.
The overview of our methods is shown in Fig.~\ref{fig:idea_overview}.

\subsection{Emphasis projection}\label{sec:emphasis projection}


One of the most straightforward approaches for minimizing the goal function $G_\z(\cdot)$ is by analyzing what trajectories that minimize $\z^* = \argmin_{\z} \Gz(\z)$ look like.
For a trajectory conditioning task, a whole trajectory $\z^*$ is directly given. Our task is to generate the rest of the motion $\x$.
With such knowledge, we can employ \textbf{imputation \& inpainting} technique by supplying the motion DPM with the $\x$-shaped $\Pzx \z^*$ to guide the generation process.

\myparagraph{Problem 1: Motion incoherence} \\
Since the imputing trajectory $\z^*$ is only a small part of the whole motion $\x$ ($L \ll N$), we often observe that the DPM ignores the change from imputation and fails to make appropriate changes on the rest of $\x$. This results in an incoherent local motion that is not aligned or well coordinated with the imputing trajectory.

\myparagraph{Solution 1: Emphasis projection} \\
We tackle this problem by giving more emphasis on the trajectory part of motion $\x$. More specifically, we propose an \textbf{Emphasis projection} method that increases the trajectory's relative importance within motion $\x$.
We achieve this by utilizing a random matrix $A = A' B$, where $A' \in \mathbb{R}^{N \times N}$ is a matrix with elements randomly sampled from $\mathcal{N}(0,1)$ and $B \in \mathbb{R}^{N \times N}$ is a diagonal matrix whose trajectory-related diagonal indexes are $c$ and the rest are $1$ for emphasizing those trajectory elements. 
In our case, we emphasize the rotation and ground location of the pelvis, $(\mathrm{rot}, x, z)$, in $\x$ by $c$ times.
We now have a projected motion $\x^\text{proj} = \frac{1}{N - 3 + 3c^2} A\x$. Note that the fractional term is to maintain the unit variance on $\x_\text{proj}$. The noising process of the projected motion becomes $q(\x^\text{proj}_t|\x^\text{proj}_0) = \mathcal{N}(\sqrt{\alpha_t}\x^\text{proj}_0, (1 - \alpha_t) \I)$. 
There is no change on how a DPM that works on the projected motion $p_\theta(\x^\text{proj}_{t-1}|\x^\text{proj}_{t})$ operates and treats $\x^\text{proj}_t$.

In Section \ref{sec:project}, we show that emphasis projection is an effective way of solving the motion incoherence problem, and is shown to be substantially better than a straightforward approach of retraining a DPM with an increased loss weight on the trajectory.

\myparagraph{Imputation on the projected motion $\x^\text{proj}$.}
We have discussed imputing on the sample $\xtone$ in Eq.~\ref{eq:imputation}. Here, we introduce an imputation on $\xzero$ which modifies the DPM's belief on the final outcome $\x_{0,\theta}$ by imputing it with $\z$. We have found this technique useful in many tasks we are interested in. 

Let us define the imputation region of $\z$ on $\x$ as $\Mzx$. We obtain the imputed $\tilde{\x}_0$ from
\begin{equation}\label{eq:imputation_x0}
\tilde{\x}_0 = (1 - \Mzx) \odot \x_{0,\theta} + \Mzx \odot \underbrace{\Pzx \z^*}_{\x \text{ shaped}}
\end{equation}

Now operating on the projected motion $\x^\text{proj}$, before we can do imputation, we need to unproject it back to the original motion using $\xzero = A^{-1}\x^\text{proj}_0$, and then project the imputed $\tilde{\x}_0$ back using $\tilde{\x}_0^\text{proj} = A \tilde{\x}_0$. We obtain the imputed motion under emphasis projection $\tilde{\x}_0^\text{proj}$ from
\begin{equation}\label{eq:impute_proj}
    \tilde{\x}_0^\text{proj} = A\Big( (1 - \Mzx) \odot (A^{-1} \x^\text{proj}_{0,\theta}) + \Mzx \odot \Pzx \z^* \Big)
\end{equation}
Substituting $\tilde{\x}_0^\text{proj}$ into Eq. \ref{eq:sampling}, we obtain the new mean $\tilde{\mu}_t^\text{proj}$ for sampling $\x_{t-1}^\text{proj} \sim \mathcal{N}(\tilde{\mu}^\text{proj}_t, \Sigma_t)$.

\subsection{Dense guidance signal with a learned denoiser}\label{sec:dense_guidance}

Another way to minimize the goal function $\Gz(\cdot)$ is by adjusting the sample of each diffusion step $\x_{t-1}$ toward a region with lower $\Gz$. This trick is called classifier guidance \cite{Dhariwal2021-bt}.
The direction of change corresponds to a score function $\nabla_\xt \log p\big(\Gx(X_t) = 0|\xt \big)$ which can be approximated as a direction $\Delta_\xzero = - \nabla_\xzero \Gz(\Pxz \x_{0,\theta})$ that reduces the goal function. We can guide the generative process by nudging the DPM's prediction as $\xzero = \x_{0,\theta} + \Delta_\xzero$. 
While imputation requires the minimizer $\z^*$ of $\Gz$, which might not be easy to obtain or may not be unique, this trick only requires the easier-to-obtain direction of change.

\myparagraph{Problem 2: Sparse guidance signal} \\
%
In the motion domain, conditioning signals can often be sparse. There are two types of sparsity that can occur: sparsity in feature and sparsity in time. \textbf{Sparsity in feature} is when the conditioning signal is a small part of the feature dimension of $\x$. For example, in trajectory-conditioned generation, $\z$ may only consist of a sequence of ground locations over time. This type of sparsity can be addressed by emphasis projection, as explained in Section \ref{sec:emphasis projection}.
\textbf{Sparsity in time} refers to cases where the conditioning signal consists of small segments of a trajectory spread out over time. 
For instance, in keyframe location conditioning task, only a sparse set of keyframe locations are given. 
When the conditioning signal-to-noise ratio becomes too small, the conditioning signal may be mistaken as noise and ignored during the denoising process.

\myparagraph{Solution 2: Dense signal propagation} \\
To turn a sparse signal into a dense signal, we need domain knowledge. One way to achieve this is by using a denoising function $f(\xt) = \xzero$, 
which is trained on a motion dataset to denoise by gathering information from the nearby motion frames. With the ability to relate a single frame to many other frames, the denoising function is capable of expanding a sparse signal into a denser one. 

We can use backward propagation through the denoising function $f$ to take advantage of this. Therefore, a dense classifier guidance can be obtained as follows:
\begin{equation}
\nabla_\xt \log p\big(\Gx(X_t) = 0|\xt \big) \approx - \nabla_\xt \Gz\big( \underbrace{\Pxz f(\xt)}_{\z \text{ shaped}} \big)
\end{equation}
While an external function can be used as $f$, we observe that the existing DPM model $\x_{0,\theta}(\xt)$ itself is a motion denoiser, and thus can be used to turn a sparse signal into a dense signal without the need for an additional model. In practice, this process amounts to computing the gradient of $G$ with respect to $\xt$ through $\x_{0,\theta}(\xt)$ using autodiff.

\myparagraph{Applying classifier guidance together with imputation.}
Whenever available, we want to utilize signals from both imputation and classifier guidance techniques to help guide the generative process.
Imputation is explicit but may encounter sparsity in time, while classifier guidance is indirect but dense. 
We want to use the direct signal from imputation wherever available (with mask $\Mzx$), and the rest from classifier guidance (with mask $1 - \Mzx$). 
Based on Eq.~\ref{eq:class_guidance}, imputation-aware classifier guidance can be written as
\begin{equation}\label{eq:class_guidance_mask}
    \mathbf{\mu}_{t} = \tilde{\mu}_{t} - (1-\Mzx) \odot s \Sigma_t \nabla_\xt \Gz\big(\Pxz f(\xt) \big)
\end{equation}
where $\tilde{\mu}$ is an imputed sampling mean. By replacing $\tilde{\mu}$ with $\tilde{\mu}^\text{proj}$, we get classifier guidance together with imputation that works with emphasis projection as
\begin{align}\label{eq:class_guidance_mask_proj}
    \Delta_\mu &= - s \Sigma_t A^{-1} \nabla_{\x_t^\text{proj}} \Gz\big(\Pxz A^{-1} f(\xt^\text{proj}) \big) \\
    \mathbf{\mu}_{t}^\text{proj} &= \tilde{\mu}_{t}^\text{proj} + A (1-\Mzx) \odot \Delta_\mu
\end{align}

\begin{figure}
    \centering
    \includegraphics[width=\linewidth]{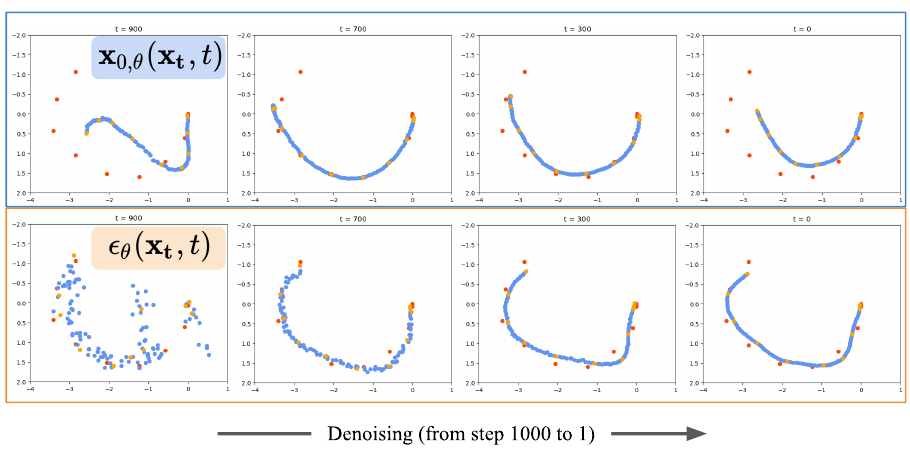}
    \caption{Comparing the evolution of the clean trajectory subject to classifier guidance from $\xzero$ and $\epsilon$ DPMs. The $\xzero$ DPM shows significant resistance on the guidance signal as exhibited by the trajectory ``contraction" behavior at $t \rightarrow 0$.}
    \label{fig:x_eps_share}
    \vspace{-1em}
\end{figure}

\myparagraph{Problem 3: DPM's bias hinders the guidance signal} \\
A DPM removes noise from an input based on the distribution of the training data it has seen. 
This could be problematic when it comes to conditional generation because the conditioning signal may be outside of the training distribution. As a result, any changes made to the classifier guidance may be reversed by the DPM in the next time step, due to its inherent bias towards the data, shown in Figure \ref{fig:x_eps_share}.

\myparagraph{Solution 3: Epsilon modeling} \\
While it is unlikely to train an unbiased DPM model, there are ways to minimize the influence of model's bias under the guidance signal. Conceptually, the DPM model usually makes less and less change near the final outcome. This is in tandem with the guidance signal that gradually decreases over time due to $\Sigma_t$ (Eq.~\ref{eq:class_guidance}).

We investigate the coefficient $\frac{\sqrt{\alpha_{t-1}} \beta_t}{1-\alpha_t}$ of $\xzero$ in the sampling mean $\mu_t$ (Eq.~\ref{eq:sampling}). This coefficient reaches its maximum value at $t=0$, meaning that an $\x_{0,\theta}$ model could have a significant impact on the sampling mean even at $t=0$, which contradicts the weak guidance signal at that time.

On the other hand, an $\epsilon_{\theta}$ model will have the most influence on the sampling mean at $t=T$, which aligns with our intuition. In Section \ref{sec:keypoint} and Figure \ref{fig:x_eps_share}, we demonstrate that modeling $\epsilon_{\theta}$ instead of $\x_{0,\theta}$ is a successful approach for managing the bias effect of the DPM model in classifier guidance. We further discuss this point in Supplementary.

\section{Applications}

\subsection{Trajectory-conditioned generation}\label{sec:traj_cond}

This task aims at generating a realistic motion $\x$ that matches a given trajectory $\z$. Our objective is to minimize the distance between the generated motion and the given trajectory, which we define as
\begin{equation}
\Gx(\x) \defeq \Big\lVert \z - \underbrace{\Pxz \x}_{\z \text{ part of } \x} \Big\lVert_p
\end{equation}
Despite the apparent simplicity of this task, a traditional DPM faces the challenge of ensuring coherence in the generated motion. However, our emphasis projection method can effectively address this problem.

\subsection{Keyframe-conditioned generation}\label{sec:keypoint_cond}
The locations of ground positions at specific times can be used to define locations that we wish the generated motion to reach. This task is a generalized version of the trajectory-conditioned generation where only a partial and potentially sparse trajectory is given.
Let $\y \in \mathbb{R}^{2 \times M}$ be a trajectory describing keyframe locations and a mask $\Myz$ describe the key motion steps.
 Our goal function of a motion $\x$ is
\begin{equation}\label{eq:goal_traj}
    \Gx(\x) \defeq \sum\nolimits_i \Big\lVert \Myz (\Pxz \x - \y) \Big\lVert_p
\end{equation}
Consequently, $\Gz(\z) = \sum_i \norm{ \Myz (\z - \y) }_p$.
Due to the partial trajectory $\y$, the imputation region of $\y$ on $\x$ becomes $\Myx = \Pzx \Myz$. 

\myparagraph{Two-stage guided motion generation.}
Generating both the trajectory and motion simultaneously under a conditioning signal can be challenging and may result in lower quality motion. To address this issue, we propose a two-step approach. First, we generate a trajectory $\z$ that satisfies the keyframe locations and then generate the motion $\x$ given the trajectory (following Section \ref{sec:traj_cond}). Our overall pipeline is depicted in Figure \ref{fig:pipeline} (a). We offer two options for generating the trajectory from keyframe locations $\y$: a point-to-point trajectory and a trajectory DPM.

The \textbf{point-to-point trajectory} connects consecutive keyframe locations with a straight line. These unrealistic trajectories can be used as imputation signals for the motion DPM during the early phase ($t \geq \tau$). 
If $\tau$ is large enough, the DPM will adjust the given trajectory to a higher quality one. However, if $\tau$ is too large, the DPM may generate a motion that does not perform well on $\Gz$.

The \textbf{trajectory DPM} $p_\phi(\z_{t-1}|\z_t)$, which is trained using the same dataset but with a smaller network, can be used to generate the trajectory under the guidance signal from $\Gz$. 
We summarize our two-stage approach in Algorithm \ref{alg:key_loc}.

It is also possible to combine the two methods, as the point-to-point trajectory can serve as a useful guidance signal for the trajectory DPM during $t \geq \tau$. After that, the trajectory DPM is subject to the usual imputation and classifier guidance from $\Gz$. By tuning $\tau$, we can balance between trajectory diversity and lower scores on $\Gz$.

\subsection{Obstacle avoidance motion generation}\label{sec:obs_cond}

Humans have the ability to navigate around obstacles while traveling from point A to B. Under our framework, this problem can be defined using two goal functions: one that navigates from A to B, called $\Gx^\text{loc}$ (defined as in Eq. \ref{eq:goal_traj}), and another that pushes back when the human model crosses the obstacle's boundary, called $\Gx^\text{obs}$, as follows
\begin{equation}
    \Gx^\text{obs}(\x) \defeq \sum_i -\mathrm{clipmax}(\mathrm{SDF}((\Pxz\x)^{(i)} ), c)
\end{equation}
where $c$ is the safe distance from the obstacle.
These two goal functions are combined additively to obtain the final goal function, $\Gx(\x) = \Gx^\text{loc}(\x) + \Gx^\text{obs}(\x)$, for this task.

We utilize the same pipeline as in Section \ref{sec:keypoint_cond}, with the exception that imputation is not possible for obstacle avoidance. Therefore, minimizing the obstacle avoidance goal relies solely on classifier guidance. 

\section{Experiments}
To evaluate our methods, we perform experiments on the standard human motion generation task conditioned on text descriptors and spatial objectives.
In particular, we evaluate (1) the performance of our model in standard text-condition motion generation tasks,
(2) the effect of emphasis projection to alleviate incoherence between spatial locations and local poses,
(3) the ability to conditionally generate motion based on spatial information by conditioning with given trajectories, keyframe locations, and obstacles.

\subsection{Settings}

\myparagraph{Evaluation metrics.}
We evaluate generative text-to-motion models using standard metrics introduced by Guo et al. \cite{guo2022t2m}. These include Fréchet Inception Distance (FID), R-Precision, and Diversity. \textbf{FID} measures the distance between the distributions of ground truth and generated motion using a pretrained motion encoder. \textbf{R-Precision} evaluates the relevance of the generated motion and its text prompt, while \textbf{Diversity} measures the variability within the generated motion. We also report \textbf{Foot skating ratio}, which measures the proportion of frames in which either foot skids more than a certain distance (2.5 cm) while maintaining contact with the ground (foot height $<$ 5 cm), as a proxy for the incoherence between trajectory and human motion.

In addition, for conditional generation with keyframe locations, we use Trajectory diversity, Trajectory error, Location error, and Average error of keyframe locations. \textbf{Trajectory diversity} measures the root mean square distance of each location of each motion step from the average location of that motion step across multiple samples with the same settings. \textbf{Trajectory error} is the ratio of unsuccessful trajectories, defined as those with \textit{any} keyframe location error exceeding a threshold. \textbf{Location error} is the ratio of keyframe locations that are not reached within a threshold distance. 
\textbf{Average error} measures the mean distance between the generated motion locations and the keyframe locations measured at the keyframe motion steps.

\myparagraph{Datasets.}
We evaluate the text-to-motion generation using the HumanML3D \cite{guo2022t2m} dataset, which is a collection of text-annotate motion sequences from AMASS \cite{mahmood2019amass} and HumanAct12 \cite{guo2020action2motion} datasets. It contains 14,646 motions and 44,970 motion annotations.

\begin{table} 
\caption{Text-to-motion evaluation on the HumanML3D \cite{guo2022t2m} dataset. The right arrow $\rightarrow$ means closer to real data is better.}
\vspace{0.5em}
\footnotesize
\centering
\begin{tabular}{lcccc}
\toprule
 & FID $\downarrow$~ & \multicolumn{1}{p{1.7cm}}{\centering R-precision $\uparrow$ \\ (Top-3)} & Diversity $\rightarrow$\\ 
 \midrule
 Real & 0.002 & 0.797 & 9.503 \\ 
 \midrule
JL2P \cite{ahuja2019language2pose} & 11.02 & 0.486 & 7.676 & \\ 
Text2Gesture \cite{bhattacharya2021text2gestures} & 7.664 & 0.345 & 6.409 & \\ 
T2M \cite{guo2022t2m} & 1.067 & 0.740 & 9.188 & \\ 
MotionDiffuse \cite{zhang2022motiondiffuse} & 0.630 & \textbf{0.782} & 9.410 & \\
MDM \cite{Tevet2022-ih} & 0.556 & 0.608 & \textbf{9.446} & \\ 
MLD \cite{chen2022mld} & 0.473  & 0.772 & 9.724 & \\  
PhysDiff \cite{yuan2022physdiff} & 0.433  & 0.631 & - & \\  
\midrule
{Ours}                  & \textbf{0.212} & 0.670 & 9.440 \\ 
{Ours $\x^\text{proj}$} & 0.235 & 0.652 & 9.726 \\ 
\bottomrule
\end{tabular}
\label{table:result_unconditional}
\vspace{-1em}
\end{table}

\begin{table} 
\caption{Trajectory-conditioned motions evaluation. The ground truth trajectory is used for imputing after each diffusion step. Comparing the effect of an original $\x$ with emphasis loss functions to the emphasis projection $\x^\text{proj}$ after imputing whole trajectories after each diffusion step.}
\vspace{0.5em}
\footnotesize
\centering
\begin{tabular}{l|ll|cc}
\toprule
 Model & Space & Emphasis & FID $\downarrow$~ & \multicolumn{1}{p{1.7cm}}{\centering Foot skating $\downarrow$ \\ ratio } \\ 
 \midrule
\multirow{4}{*}{MDM\cite{Tevet2022-ih}}   & $\x$            & loss $1 \times$        & 0.904 &  0.284 \\ 
                                          \cmidrule{2-5}
                                          & \multirow{4}{*}{$\x^\text{proj}$} & $c=1$   & 0.632 &  0.304 \\ 
                                          &                                   & $c=2$   & 0.464 &  0.309 \\
                                          &                                   & $c=5$   & 0.466 &  0.256 \\
                                          &                                   & $c=10$  & 1.029 &  0.161 \\ 
\midrule
\multirow{8}{*}{Ours}   & \multirow{4}{*}{$\x$}     & loss $1 \times$                & 0.278 &  0.262 \\
                        &                           & loss $2^2 \times$              & 0.256 &  0.250 \\
                        &                           & loss $5^2 \times$              & 0.240 &  0.249 \\
                        &                           & loss $10^2 \times$             & 0.320 &  0.265 \\
                        \cmidrule{2-5}
                        & \multirow{4}{*}{$\x^\text{proj}$}     & $c=1$ & 0.307 & 0.268 \\
                        &                           & $c=2$             & 0.290 & 0.257 \\ 
                        &                           & $c=5$             & 0.274 & 0.199 \\ 
                        &                           & $c=10$            & \textbf{0.198} & \textbf{0.128} \\ 
\bottomrule
\end{tabular}
\label{table:result_projection}
\end{table}

\myparagraph{Implementation details.}
Both our motion DPM and trajectory DPM are based on UNET with AdaGN \cite{Dhariwal2021-bt} depicted in details in the Supplementary. The motion DPM is an $\xzero$ model, while the trajectory DPM is an $\epsilon$ model, as explained in Section \ref{sec:dense_guidance}, to enhance controllability. We utilized DDPM  \cite{Ho2020-ew} with $T$=1,000 denoising steps for training and inference of both models. Additionally, we condition the generation process on text prompts in a classifier-free \cite{ho2021classifierfree} manner, similar to MDM \cite{Tevet2022-ih}, and use the CLIP \cite{radford2021learning} model as the text encoder across all tasks.

\myparagraph{Computational resources}
Our \methodname~architecture is capable of running both the motion and trajectory models on a single commercial GPU, such as the Nvidia RTX 2080 Ti, 3080, or 3090. The trajectory model achieved a throughput of 2,048 samples per second when run on an RTX 3090, with a training time of approximately 4.34 GPU hours. Meanwhile, the motion model achieved a throughput of 256 samples per second on an RTX 3090, with a training time of around 34.7 GPU hours. The total inference time for one sample is approximately 110 seconds.

\subsection{Text-to-motion generation}\label{sec:text2motion}
This section evaluates our model's performance in the standard text-to-motion generation task and compares it with other motion DPM baselines: MotionDiffuse~\cite{zhang2022motiondiffuse}, MDM~\cite{Tevet2022-ih}, MLD~\cite{chen2022mld}, and PhysDiff~\cite{yuan2022physdiff}.
Tab.~\ref{table:result_unconditional} shows the results where our model architecture outperforms the baselines significantly in terms of motion quality measured by FID, while maintaining similar R-Precision and Diversity.


\begin{table*}
\caption{The effect of different conditioning strategies tested on keyframe-conditioning task. The keyframes ($N=5$) are sampled from the ground truth motion trajectories with the same text prompts in the HumanML3D \cite{guo2022t2m} test set.}
\vspace{0.5em}
\footnotesize
\centering
\begin{tabular}{l|ll|ccccccc}
\toprule
 Model & \multicolumn{2}{c|}{Conditioning} & FID $\downarrow$~ & \multicolumn{1}{p{1.6cm}}{\centering Foot $\downarrow$ \\ skating ratio } & \multicolumn{1}{p{1.6cm}}{\centering Traj. $\uparrow$ diversity (m.)} & \multicolumn{1}{p{1.3cm}}{\centering Traj. err. $\downarrow$ \\ (50 cm)} & \multicolumn{1}{p{1.3cm}}{\centering Loc. err. $\downarrow$ \\ (50 cm)} & \multicolumn{1}{p{1.3cm}}{\centering Avg. err. $\downarrow$} & \multicolumn{1}{p{1.7cm}}{\centering R-precision $\uparrow$ \\ (Top-3)} \\ \midrule
\multirow{4}{*}{MDM\cite{Tevet2022-ih}} & \multirow{4}{*}{\shortstack{Single \\ stage}} & $\x$ + $\tau$=0   & 1.256 & 0.202 & 0.134 & 0.000 & 0.000 & 0.000 & 0.631 \\ 
                                        &        & $\x^\text{proj}$ + $\tau$=0              & 2.994 & 0.151 & 0.134 & 0.000 & 0.000 & 0.000 & 0.554 \\ 
                                        &        & $\x^\text{proj}$ + $\tau$=100            & 2.213 & 0.095 & 0.214 & 0.326 & 0.127 & 0.236 & 0.555 \\ 
                                        &        & $\x^\text{proj}$ + no p2p                & 1.679 & 0.092 & \textbf{0.394} & 0.519 & 0.326 & 0.543 & 0.548 \\ 
\midrule
\multirow{8}{*}{Ours ($\x^\text{proj}$)}   & \multirow{2}{*}{\shortstack{Single \\ stage}} & $\tau$=0       & 0.902 & 0.127 & 0.117 & 0.000 & 0.000 & 0.000 & 0.594 \\ 
                        &                               & $\tau$=100                        & \textbf{0.523} & \textbf{0.086} & 0.157 & 0.176 & 0.049 & 0.139 & 0.599 \\ \cmidrule{2-10}
                        & \multirow{5}{*}{\shortstack{Two \\ stage}}    & $\tau$=100                        & 0.937 & 0.098 & 0.120 & 0.076 & 0.020 & 0.109 & 0.574 \\ 
                        &                               & $\tau$=300                        & 0.938 & 0.098 & 0.127 & 0.118 & 0.031 & 0.128 & 0.573 \\ 
                        &                               & $\tau$=500                        & 0.908 & 0.098 & 0.140 & 0.157 & 0.043 & 0.140 & 0.577 \\ 
                        &                               & $\tau$=700                        & 0.898 & 0.098 & 0.162 & 0.196 & 0.058 & 0.153 & 0.580 \\ 
                        &                               & $\tau$=900                        & 0.874 & 0.098 & 0.192 & 0.238 & 0.080 & 0.180 & 0.581 \\ 
                        &                               & no p2p                            & 0.862 & 0.104 & 0.222 & 0.287 & 0.118 & 0.282 & 0.577 \\
\bottomrule
\end{tabular}
\label{table:result_kps}
\vspace{-1em}
\end{table*}

\begin{figure}
    \centering
    \includegraphics[width=\linewidth]{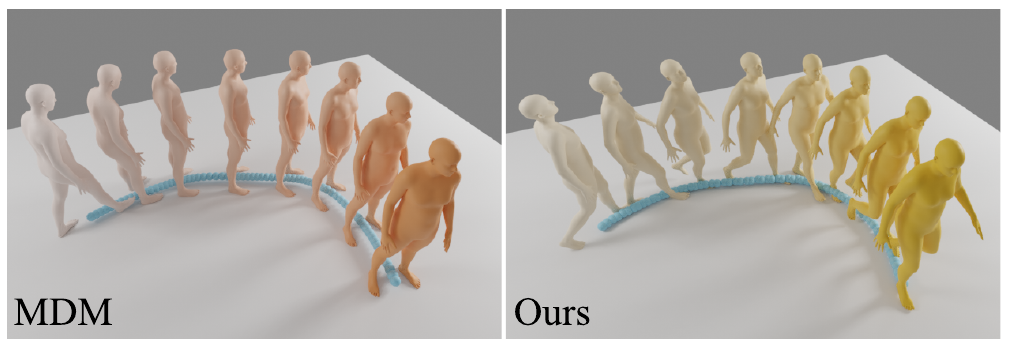}
    \caption{Generated motion, conditioned a given trajectory and text ``walking forward". 
    MDM \cite{Tevet2022-ih} exhibits motion incoherence where the model disregards the trajectory and generates an inconsistent motion. Our method, improved by emphasis projection, deals effectively with the conditioning.}
    \label{fig:incoherence}
\end{figure}

\subsection{Trajectory-conditioned generation}\label{sec:project}

This section demonstrates how our emphasis projection method can address the issue of incoherent motion caused by spatial conditioning, specifically in the trajectory conditioning task, where the model is provided with ground-truth trajectories for imputation at each denoising step and is required to generate corresponding local poses.
Both quantitative and qualitative results support that our emphasis projection leads to a reduction in Foot skating ratio, as evidenced in Tab.~\ref{table:result_projection} and a more coherent motion in Fig.~\ref{fig:incoherence} compared to the MDM \cite{Tevet2022-ih} model.

We also compare our emphasis projection method with an alternative approach of increasing the trajectory loss strength during training. We include $\text{loss } k^2\times$ baselines, where $k \in \{1,2,5,10\}$, for comparison. The results in Tab.~\ref{table:result_projection} indicate that, while increasing the loss strength marginally improves both FID and Foot skating ratio, increasing it beyond a certain point leads to a decline in both FID and Foot skating ratio. By contrast, our emphasis projection method consistently leads to improvements in both metrics. 
We discuss this topic further in the Supplementary.


\begin{figure}
    \centering
    \includegraphics[width=\linewidth]{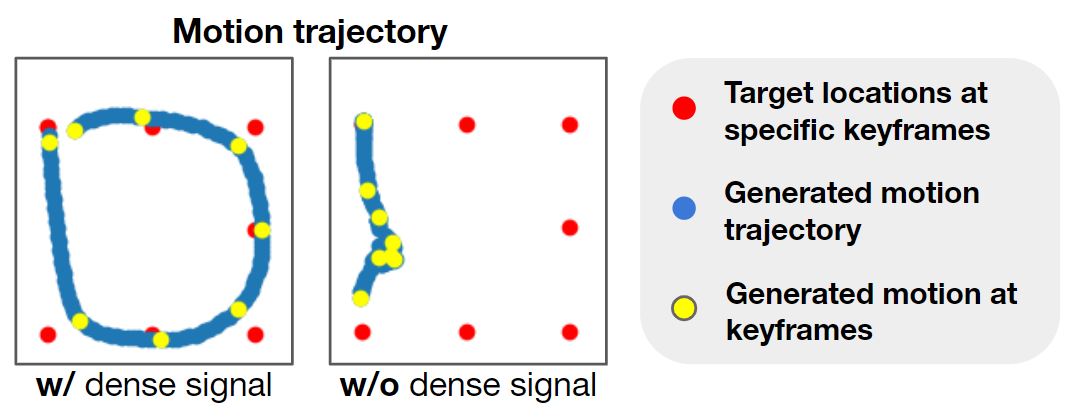}
    \caption{Generated motion trajectories, conditioned on target locations at given keyframes.
    Without dense signal propagation, the model ignores the target conditions.}
    \label{fig:dense_signal}
\end{figure}

\subsection{Keyframe-conditioned generation}\label{sec:keypoint}

This section evaluates the quality and adherence of the generated motion to the desired goal. A viable solution must meet both criteria to an acceptable degree.

To achieve high-quality motion, both FID and Foot skating ratio are essential since FID alone cannot adequately measure the trajectory-motion coherence.
Our Emphasis projection technique significantly improves motion coherence, reducing foot skating as shown in Tab.~\ref{table:result_kps} while MDM \cite{Tevet2022-ih} is unsuitable for this task due to the high motion incoherence. Furthermore, our improved architecture significantly improves motion quality in all cases.
Note that without dense signal propagation, the model ignores the keyframe conditioning as shown in Fig~\ref{fig:dense_signal}.

While a single-stage model performs reasonably well due to emphasis projection, it is too restrictive at $\tau=0$ (forced trajectory), resulting in relatively high Foot skating. This issue can be addressed by allowing more modification (increasing to $\tau$ to $100$) but at the cost of higher Loc. error.

Lastly, the trajectory model's better controllability reduces Location error by more than half compared to the single-stage model at $\tau=100$. As expected, increasing $\tau$ leads to more freedom in the model, resulting in increased Trajectory diversity, lower FID, and higher Location error.

\subsection{Obstacle avoidance motion generation}\label{sec:avoidance}
\begin{figure}
    \centering
    \includegraphics[width=\linewidth]{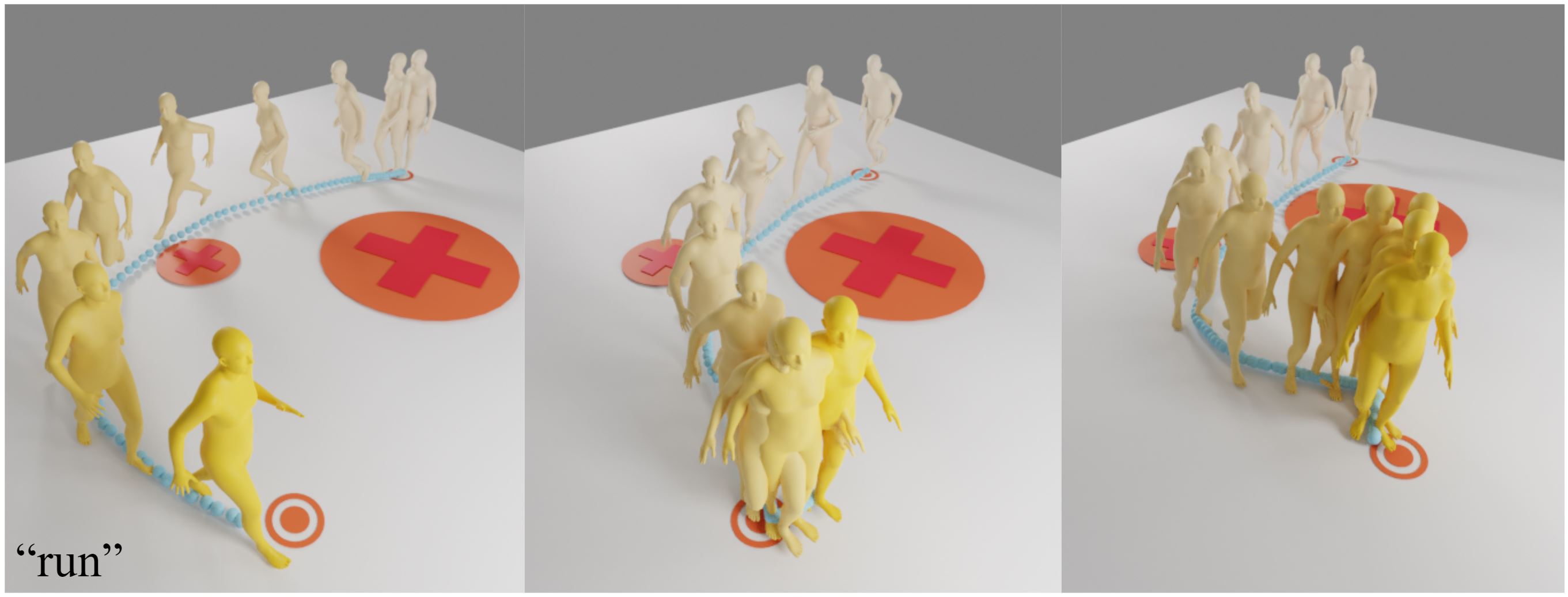}
    \caption{Qualitative results from the obstacle avoidance task given keyframe locations and obstacles. The red crossed areas represent obstacles to avoid. More results are in the supplementary.}
    \label{fig:collision}
\end{figure}

Finally, we demonstrate our model's ability to generate motion on additional guidance on the obstacle avoidance task.
In this task, we randomly sample the target point that the human needs to reach at a specific motion step along with a set of obstacles it needs to avoid, represented as a 2D SDF (Sec.~\ref{sec:obs_cond}). We show the qualitative results in Fig~\ref{fig:collision}.

\section{Discussion and Limitations}
In this work, we propose \methodname, a controllable human motion generation method based on goal functions. \methodname~produces high-quality and diverse motions and supports diverse possibilities for goal functions. 
Since obtaining necessary data and designing a classifier-free learning method for non-explicit goals, such as obstacle avoidance, can be challenging, our \methodname~utilizes a classifier-based method which allows for more conditioning flexibility without retraining the model.  
Thus, our studies on effective classifier guidance will be useful for further including more guiding signals.

\myparagraph{Acknowledgement.}
This work was supported by the SNSF project grant 200021 204840.



{\small
\bibliographystyle{ieee_fullname}
\bibliography{reference}
}

\newpage
\clearpage

\begingroup

\appendix
\twocolumn[
\begin{center}
\Large{\bf Guided Motion Diffusion for Controllable Human Motion Synthesis \\ **Appendix**}
\end{center}
]

\counterwithin{table}{section}
\counterwithin{figure}{section}
\setcounter{page}{1}

\section{Analysis on $\mathbf{x}_{0,\theta}$ vs. $\epsilon_{\theta}$ DPMs}
\label{sec:analysis_xzero_vs_epsilon}

In this section, we discuss the differences in behavior between the $\mathbf{x}_{0,\theta}$ and $\epsilon_{\theta}$ models used to train DPMs. While both models are capable of generating high-quality samples, their denoising processes differ significantly. 
In Section 4.2, we previously claimed that the $\xzero$ predicting model maximizes its influence on the outcome $\xtone$ when $t \rightarrow 0$, whereas the $\epsilon$ predicting model maximizes its influence when $t \rightarrow T$. Based on this observation, we argue that the $\epsilon$ predicting model is more favorable than the $\xzero$ predicting model in circumstances where the outcome of the diffusion process will be altered by an external factor from the classifier.

To further understand the behavior of the two models, we examine Equation 1, which indicates that $\xtone$ is sampled from a Normal distribution with mean 
\begin{equation}
    \mu_t = \underbrace{\frac{\sqrt{\alpha_{t-1}}\beta_t}{1 - \alpha_t}}_{a} \xzero + \underbrace{\frac{\sqrt{1-\beta_t} (1 - \alpha_{t-1})}{1-\alpha_t}}_{b} \xt
\end{equation}
The coefficients $a$ and $b$ in $\mu_t$ modulate the contribution of the $\xzero$ model and the previous output $\xt$. The larger the coefficient $a$ is relative to $b$, the larger the contribution of the $\xzero$ model on the outcome of the denoising process. 

In the case of an $\epsilon$ model, we substitute $\xzero$ based on the relationship $\xzero = \frac{\xt - \sqrt{1-\alpha_t} \epsilon}{\sqrt{\alpha_t}}$ and get a different expression for $\mu_t$ as
\begin{equation}
    \mu_t = \underbrace{\left( \frac{a}{\alpha_t} + b \right)}_{c} \xt - \underbrace{\frac{a \sqrt{1-\alpha_t}}{\sqrt{\alpha_t}}}_{d} \epsilon
\end{equation}
We can see that the contribution of the $\xzero$ model and the $\epsilon$ model are starkly different, with the $\epsilon$ model having a stronger contribution on $\mu_t$, and hence $\xtone$, where $t$ is large, while the opposite is true for the $\xzero$ model. In other words, an $\epsilon$ model is restricted to make a smaller change over time while an $\xzero$ model can still make a large change even at the very end of the diffusion process.

From the analysis above, we conclude that the choice of modeling $\epsilon$ or $\xzero$ is no longer arbitrary. Given the fact that the classifier guidance strength is modulated by $\Sigma_t$, which is smaller as $t \rightarrow 0$, and the fact that all DPM models are biased toward their training datasets, an $\xzero$ model capable of ever larger change as the guidance signal diminishes is not an ideal choice because it could easily overpower the guidance signal, especially at the end of the diffusion process, undoing all the guidance signal. Therefore, our \methodname's trajectory model, which is subject to classifier guidance signals, is carefully chosen to be an $\epsilon$ model. We visualize the relative share over time of each model on $\mu_t$ in Figure \ref{fig:x_eps_share_ratio} and show the impact of the choice of model in Figure 3 in the main paper. 

\begin{figure}
    \centering
    \includegraphics[width=1\linewidth]{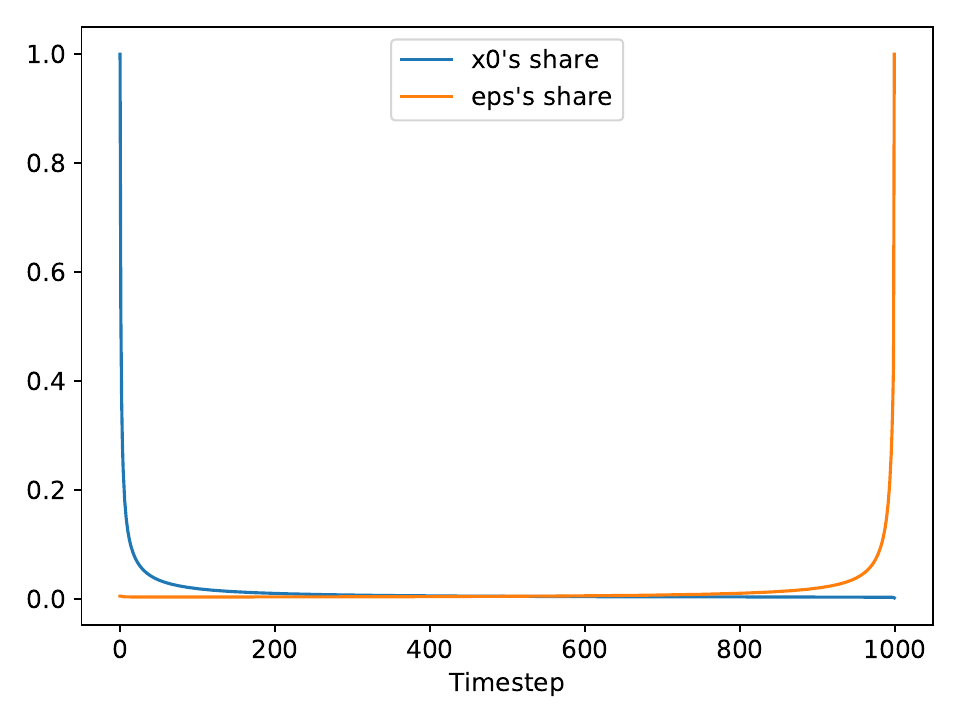}
    \caption{Comparing $\xzero$ and $\epsilon$ contributions in the prediction of $\xtone$ based on the Cosine $\beta_i$ scheduler.}
    \label{fig:x_eps_share_ratio}
\end{figure}

\subsection{Challenges of modeling $\epsilon$ in practice}

In Section \ref{sec:analysis_xzero_vs_epsilon}, we discussed the benefits of modeling $\epsilon$ over $\xzero$ from the perspective of classifier guidance. However, there are fundamental differences and requirements for architectures that excel in predicting $\xzero$ versus $\epsilon$. Specifically, $\epsilon \sim \mathcal{N}(\mathbf{0}, \mathbf{I})$ is independent and full-rank, meaning there is no smaller latent manifold that it resides in. On the other hand, $\xzero$ usually has a smaller latent manifold, which is the case for many real-world data including motions as most of the possible values in $\x \in \mathbb{R}^{263 \times M}$ are not valid human motions, only a small subset of that is. Due to these differences, it requires special considerations for architectural design in models that successfully predict $\epsilon$.

Although there is no sufficient reason to believe that modeling $\epsilon$ is fundamentally harder than modeling $\xzero$, in practice, modeling $\epsilon$ is restricted to cases where its shape is relatively small compared to the latent dimension of the denoising model. For example, when modeling $\epsilon \in \mathbb{R}^{263 \times M}$ for a motion DPM, the original MDM architecture for $\epsilon$ prediction generated low-quality jagged motions compared to the same architecture for $\xzero$ prediction, which produced high-quality motions flawlessly. Increasing the latent dimension of MDM from 512 to 1,536 did not solve the problem entirely, indicating that predicting $\xzero$ and $\epsilon$ requires different architectural designs that may not be satisfied by a single architecture. 
We argue that there require further studies on how to effectively design an $\epsilon$ predicting model.

However, when the space of $\epsilon$ is relatively small, such as in a trajectory DPM, the choice of architecture seems to matter less. The MDM transformer architecture was applied to trajectory modeling with relatively no problem. Ultimately, we used a convolution-based UNET with AdaGN \cite{Dhariwal2021-bt} as the final architecture for our proposed method, as it demonstrated superior performance for both modeling trajectories and motions.

\section{Relative vs. Absolute root representation}
In this section, we discuss different ways of representing the root locations $\z$ of the motion.
Generally, the root locations can be represented as absolute rotations and translations (\textbf{abs}) or relative rotations and translations compared to the previous frame (\textbf{rel}).
In MDM \cite{Tevet2022-ih}, the root locations are represented with relative representation following the HumanML3D \cite{guo2022t2m} dataset. In this case, the global locations at an exact frame $i$ can be obtained by a cumulative summation of rotations and translations before $i$.

However, we observe that representing the root with absolute coordinates (\textbf{abs}) is more favorable than the relative one (\textbf{rel}) in two aspects: being more straightforward for imputation and easier to optimize. Therefore, we adopt the absolute root representation for our models.

In \textbf{rel}, a trajectory is described as velocity $\Delta \z^{(i)}/\Delta i$ in the local coordinate frame of the current pelvis rotation.
This representation makes each $\z^{(i)}$ dependent on all previous motion steps in a non-linear relationship. Optimization becomes less stable as a small change in early motion steps may compound and become a larger change later on. Also, imputing specific values becomes ill-posed since there are many possible sets of values that are satisfiable. 

On the other hand, for \textbf{abs}, the imputation and optimization of $\z$ become straightforward as they only involve replacing or updating $\z^{(i)}$ without dependency on other motion steps.
We ablated the root representation by retraining MDM \cite{Tevet2022-ih} and our model with both relative and absolute root representation, then show the results in Tab~\ref{table:abs_rel_root}.
MDM shows a significant drop in performance when converted to the absolute representation, likely because the architecture is highly optimized for the relative representation, while for our models, the representation change results in a trade-off between the \textit{FID} and \textit{R-precision}.

Lastly, we note that the use of absolute root representation is necessary for our final model as the spatial guidance is done via a combination of imputation and optimization.

\begin{table} 
\caption{Text-to-motion evaluation on the HumanML3D \cite{guo2022t2m} dataset. Comparision between relative and absolute root representation. The right arrow $\rightarrow$ means closer to real data is better.}
\vspace{0.5em}
\footnotesize
\centering
\begin{tabular}{lcccc}
\toprule
 & FID $\downarrow$~ & \multicolumn{1}{p{1.7cm}}{\centering R-precision $\uparrow$ \\ (Top-3)} & Diversity $\rightarrow$\\ 
 \midrule
 Real & 0.002 & 0.797 & 9.503 \\ 
 \midrule
MDM \cite{Tevet2022-ih} (\textbf{rel})  & 0.556 & 0.608 & \textbf{9.446} & \\ 
MDM \cite{Tevet2022-ih} (\textbf{abs}) & 0.894 & 0.638 & 8.819 & \\
\midrule
{Ours} (\textbf{rel}) & 0.305 & 0.666 & 9.861 \\
{Ours} (\textbf{abs}) & \textbf{0.212} & \textbf{0.670} & 9.440 \\
\midrule
{Ours $\x^\text{proj}$} & 0.235 & 0.652 & 9.726 \\ 
\bottomrule
\end{tabular}
\label{table:abs_rel_root}
\vspace{-1em}
\end{table}

\section{Analysis on Emphasis projection}\label{supp:emphasis_proj}

In this section, we discuss in greater detail our proposed Emphasis projection. 
Conceptually, we wish to increase the relative importance of the trajectory representation $\z$ within the motion representation $\x$. 
This could be done most simply by increasing the magnitude of those values of $\z$ by multiplying it with a constant $c > 1$. 
More precisely, let us assume the shape of $x$ is $263 \times M$. A single motion frame $x = \x^{(i)}$ is a column vector of $263$ scalars in which $3$ elements $(\mathrm{rot}, x, z)$ are a column vector of a trajectory frame $z = \z^{(i)}$ that comprises root rotation and a ground location. The new trajectory elements become $z \times c$.

\myparagraph{How to calculate a suitable scalar $c$?}

By introducing a scalar $c>1$, the trajectory elements $z$ are given a higher relative importance than the remaining $260$ elements in $x$. This relative importance is determined by the cumulative variance of the $z$ elements compared to that of the remaining $260$ elements. Assuming that all elements in $x$ are independently and identically distributed according to a standard Normal distribution $\mathcal{N}(0,1)$, we can represent the cumulative variance of the trajectory elements as
\begin{equation}
    \mathrm{Var}[x^{(\mathrm{rot})} + x^{(\mathrm{x})} + x^{(\mathrm{z})}] = \sum_{j \in \text{Traj.}} \mathrm{Var}[x^{(j)}] = 3
\end{equation} 
where $j \in \text{Traj.}$ refers to the indexes in $x$ that are related to trajectory.

Similarly, we can represent the cumulative variance of the remaining $260$ elements as $\mathrm{Var}[\sum_{j \notin \text{Traj.}} x^{(j)}] = 260$, where $j \notin \text{Traj.}$ refers to the indexes in $x$ that are not related to trajectory. 

When we multiply trajectory by $c$, the new cumulative variance becomes $\mathrm{Var}[c \times (x^{(\mathrm{rot})} + x^{(\mathrm{x})} + x^{(\mathrm{z})})] = c^2 \sum_{j \in \text{Traj.} } \mathrm{Var}[x^{(j)}] = 3 c^2$. Therefore, the relative importance of the scaled trajectory elements compared to the remaining $260$ elements in $x$ is given by the expression 
\begin{equation}
    \frac{3 c^2}{260 + 3c^2}
\end{equation}
Setting $c = \sqrt{\frac{260}{3}} \approx 9.3$ results in a relative importance of 50\%, which strikes a reasonable balance between the trajectory and the rest of human motion. We have selected $c = 10$ as a rounded number of this fact, and it has been found to work well in practice.

\myparagraph{Maintaining the uniform unit variance after scaling}

After scaling up the trajectory elements by a factor of $c$, the variance of the new motion representation is no longer uniform. This presents a problem when trying to model it using the original DPM's $\beta_t$ scheduler. In order to maintain uniform variance, we can redistribute the increased values from the trajectory part $c \times z$ to the rest in $x$ via a random matrix projection.

There are two reasons why a random matrix projection is a good choice. First, it maintains the distance measure of the original space with high probability, meaning that the properties of the motion representation remain relatively unchanged. Second, a random matrix projection is easy to obtain and linear. It has an exact inverse projection, which ensures that there is no loss of information after the projection.

Finally, to maintain unit variance, we scale down the entire vector uniformly by a factor of $\frac{1}{263 - 3 + 3c^2}$.

\subsection{Trajectory loss scaling}\label{supp:loss_scaling}

One approach to increase the emphasis on the trajectory part $\z^{(i)}$ of the motion $\x^{(i)}$ is to scale the reconstruction loss of only the trajectory part during the training of the motion DPM. This method does not change the representation but can potentially increase the model's emphasis on the trajectory part of the motion compared to the rest of the motion.

To compare the loss scaling method with the proposed Emphasis projection, we formulate a new loss function for a specific motion frame $i$, which increases the trajectory importance by a factor of $k$. This is given by the equation:
\begin{equation}
\mathcal{L}^{(i)}_k = \sum_{j \in \text{Traj.}} \norm{ k \hat{x}^{(j)} - k x^{(j)} }^2 + \sum_{j \notin \text{Traj.}} \norm{ \hat{x}^{(j)} - x^{(j)} }^2
\end{equation}
Here, $\hat{x} = \x_{0, \theta}(\xt)^{(i)}$ represents the $i$-th motion frame of the DPM's prediction and $x = \xzero^{(i)}$ represents the $i$-th motion frame of the ground truth motion. The value of $k$ multiplies inside the squared loss, resulting in $k^2$ times more importance on the trajectory part of the motion. For example, setting $k=10$ would increase the importance of the trajectory part by 100-fold, which has the same scaling effect as setting $c=10$ in Emphasis projection. Hence, the reasonable range of $k$ is the same as that of $c$.

In the main text, we experimented with $k \in { 1,2,5,10 }$ and found that Emphasis projection consistently outperformed loss scaling regarding motion coherence.

\section{\methodname's Model Architecture}

The trajectory and motion architectures of \methodname are both based on UNET with Adaptive Group Normalization (AdaGN), which was originally proposed by \cite{Dhariwal2021-bt} for class-conditional image generation tasks. However, we have adapted this model for sequential prediction tasks by using 1D convolutions. It should be noted that our architectures share some similarities with \cite{Janner2022-no} with the addition of AdaGN. The architecture overview is depicted in Figure \ref{fig:arch1} while the Adaptive Group Normalization is depicted in Figure \ref{fig:arch2}. The hyperparameter settings of the two DPMs are shown in Table \ref{tab:arch}. We currently are in the process of open-sourcing the code base of \methodname.

Convolution-based architectures are commonly used in state-of-the-art image-domain DPMs, such as those proposed by \cite{Rombach2021-nu} and \cite{Saharia2022-ns}. On the other hand, transformer-based architectures, which were used in the original MDM proposed by \cite{Tevet2022-ih}, are not well-studied architectures for DPMs \cite{Cao2022-th, Peebles2022-nw}.

Our proposed architecture alone has led to a significant improvement in motion generation tasks, reducing the Fréchet Inception Distance (FID) by more than half compared to the original MDM (0.556 vs 0.212), as shown in Table 1 in the main paper.

\begin{table}[h]
\caption{Network architecture of our \methodname's models based on the proposed 1D UNET with AdaGN.}
\label{tab:arch}
\begin{center}
\begin{tabular}{l|cc}
\toprule
\textbf{Parameter}           & \textbf{Trajectory DPM}  & \textbf{Motion DPM}   \\
\midrule
Batch size                   & 512                 & 64               \\
Base channels                & 512                 & 512                \\
Channel multipliers          & {[}0.125, 0.25, 0.5{]}        &  {[}2, 2, 2, 2{]}    \\
Attention resolution         & No attention             & No attention          \\
Samples trained              & \multicolumn{2}{c}{32M}              \\
$\beta$ scheduler            & \multicolumn{2}{c}{Consine \cite{Nichol2021-vw}}            \\
Learning rate                & \multicolumn{2}{c}{1e-4}                 \\
Optimizer                    & \multicolumn{2}{c}{AdamW (wd = 1e-2)}  \\
Training $T$                 & \multicolumn{2}{c}{1000}                 \\
Diffusion loss               & $\epsilon$ prediction & $\xzero$ prediction \\
Diffusion var.               & \multicolumn{2}{c}{Fixed small $\tilde{\beta}_t = \frac{1-\alpha_{t-1}}{1-\alpha_t} \beta_t$} \\       
Model avg. beta               & \multicolumn{2}{c}{0.9999} \\       
\bottomrule
\end{tabular}
\end{center}
\end{table}

\begin{figure}
    \centering
    \includegraphics[width=1.0\linewidth]{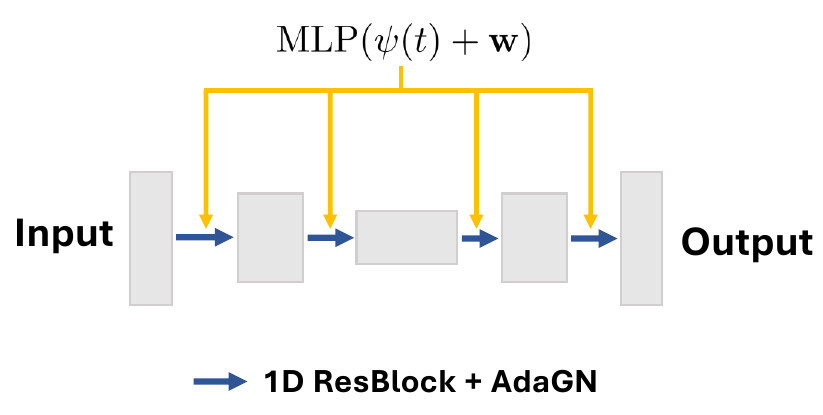}
    \caption{A simplified overview of our \methodname's 1D UNET + AdaGN architecture that is designed to process two input signals: the time step $\psi(t)$ and a text-prompt embedding $\textbf{w}$. The time step is encoded using sinusoidal functions, while the text-prompt embedding is generated by the CLIP text encoder model, as described in \cite{Tevet2022-ih}. The ResBlock + AdaGN component of the model is explained in Figure \ref{fig:arch2}.}
    \label{fig:arch1}
\end{figure}

\begin{figure}
    \centering
    \includegraphics[width=0.8\linewidth]{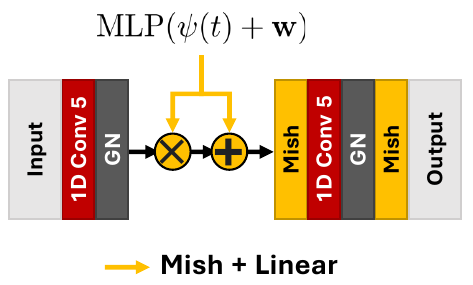}
    \caption{A single 1D ResBlock with Adaptive Group normalization (AdaGN) \cite{Dhariwal2021-bt}. The conditioning signal from the MLP, shared across all ResBlocks, is projected by first applying a Mish activation and then a resizing linear projection specific to each ResBlock. All kernel sizes are 5. We use Mish activation function following \cite{Janner2022-no}.}
    \label{fig:arch2}
\end{figure}





\section{Training details}

\textbf{\methodname's models.} We used a batch size of 64 for motion models and a batch size of 512 for trajectory models. 
No dropout was used in all of the \methodname's models: both trajectory and motion. 
We used AdamW with a learning rate of 0.0001 and weight decay of 0.01. We clipped the gradient norm to 1 which was found to increase training stability. We used mixed precision during training and inference.
We trained all motion models for 32,000,000 samples (equivalent to 500,000 iterations at batch size 64, and 62,500 iterations at batch size 512). 
We also employed the moving average of models during training ($\beta = 0.9999$) \cite{Ho2020-ew} and used the averaged model for better generation quality. Do note that our model architecture still improves over the baseline MDM without the moving average.

\textbf{\methodname's trajectory model.}
While the crucial trajectory elements are only the ground x-z locations, we have found it useful to train the trajectory model with all four components (rot, x, y, z). The additional (rot, y) seem to provide useful information that helps the model learn and reduce overfitting in the trajectory model.
Note that the trajectory DPM is sensitive to the overfitting problem. Overtraining the model will result in a strong trajectory bias in the model making the model more resistant to classifier guidance and imputation. Our choice of training the trajectory model for 32,000,000 samples was carefully chosen based on this observation.

\textbf{Retraining of MDM models.} 
We retrained the original MDM using our absolute root representation and proposed Emphasis projection as the two main baselines. In order to maintain consistency, we kept the original optimization settings for the MDM models. Specifically, we used AdamW optimizer with a learning rate of 0.0001 and without weight decay. We found that gradient clipping of 1 provided more stability, so we also applied it here. We did not utilize mixed precision training for these models. To match the settings of the original MDM, we trained these models for 400,000 iterations at a batch size of 64.

\section{Inferencing details}

We have chosen the value of $s$ as $100$ for the classifier guidance strength. Our experiments have shown that this value of $s$ performs well within the range of $100$ to $200$. For all our goal functions $\Gx$, we always used the $p=1$ norm. Whenever feasible, we implemented both imputation and classifier guidance concurrently. However, we ceased the guidance signals, i.e., classifier guidance and imputation, at $t=20$ as this led to a slight improvement in the motion coherence.

\myparagraph{Obstacle avoidance task.} In this particular case, it was not feasible to create a point-to-point trajectory because doing so could potentially lead to a collision with an obstacle. As a result, we decided against utilizing any point-to-point trajectory imputation for this task.


\end{document}